\documentclass[runningheads]{llncs}

\usepackage[T1]{fontenc}
\usepackage{graphicx}
\usepackage{hyperref}
\usepackage{url}

\usepackage{xcolor}
\usepackage{ulem}

\usepackage{tikz}
\usetikzlibrary{arrows.meta,positioning}

\definecolor{edgepink}{RGB}{176,62,116}
\definecolor{softpink}{RGB}{252,238,244}
\definecolor{softgray}{RGB}{246,248,252}
\definecolor{arrowgray}{RGB}{90,90,95}

\usepackage{booktabs,tabularx,makecell,ragged2e,float}
\newcolumntype{Y}{>{\RaggedRight\arraybackslash}X}

\usepackage[most]{tcolorbox}
\usepackage{enumitem}
\usepackage{placeins}
\usepackage{needspace}
\setlist{topsep=2pt,itemsep=2pt,parsep=0pt,partopsep=0pt}
\newcommand{\relpar}[1]{%
  \vspace{0.75em}%
  \noindent\textbf{#1}\par\vspace{0.25em}%
}

\begin{document}
\title{IC-EO: Interpretable Code-based assistant for Earth Observation}
\author{Lamia Lahouel*\inst{1} \and
Laurynas Lopata*\inst{2} \and
Simon Gruening\inst{2} \and
Gabriele Meoni\inst{3,4} \and
Gaetan Petit\inst{2}\and
Sylvain Lobry\inst{1}\orcidID{0000-0003-4738-2416}}
\authorrunning{L. Lahouel et al.}
\institute{Université Paris Cité, LIPADE, F-75006 Paris, France \and
askEarth AG, Zurich, Switzerland \and
ESA $\Phi$-lab, Frascati, Italy \and 
ESA, Advanced Concepts and Studies Office, Noordwijk, the Netherlands}
\maketitle              % typeset the header of the contribution
\begin{abstract}
Despite recent advances in computer vision, Earth Observation (EO) analysis remains difficult to perform for the laymen, requiring expert knowledge and technical capabilities. Furthermore, many systems return black-box predictions that are difficult to audit or reproduce. Leveraging recent advances in tool LLMs, this study proposes a conversational, code-generating agent that transforms natural-language queries into executable, auditable Python workflows. The agent operates over a unified easily extendable API for classification, segmentation, detection (oriented bounding boxes), spectral indices, and geospatial operators. With our proposed framework, it is possible to control the results at three levels: (i) tool-level performance on public EO benchmarks; (ii) at the agent-level to understand the capacity to generate valid, hallucination-free code; and (iii) at the task-level on specific use cases. In this work, we select two use-cases of interest: land-composition mapping and post-wildfire damage assessment. The proposed agent outperforms general-purpose LLM/VLM baselines (GPT-4o, LLaVA), achieving 64.2\% vs.\ 51.7\% accuracy on land-composition and 50\% vs.\ 0\% on post-wildfire analysis, while producing results that are transparent and easy to interpret. By outputting verifiable code, the approach turns EO analysis into a transparent, reproducible process.
\keywords{Earth Observation \and Large language models \and Tool-using agents \and Geospatial analysis}
\end{abstract}

\section{Introduction}
\label{sec:intro}
Earth Observation (EO) has evolved into a data-rich field with applications in a wide-range of problematics such as environmental monitoring, disaster response, agriculture, and urban planning. However, EO users face fragmented toolchains: imagery retrieval, cloud masking, reprojection and tiling, model selection, and geodesic statistics live across different platforms and libraries, with many end-to-end pipelines yielding black-box outputs that are hard to trust or reproduce \cite{gorelick2017earthengine,stewart2022torchgeo}. This fragmentation raises the barrier for non-experts and operational use.

Concurrently, large language models (LLMs) have progressed from text generators into tool-using controllers capable of planning, calling APIs, and writing executable programs. The shift from transformer foundations to instruction following and code centric reasoning suggests a path to unify complex workflows while keeping each step explicit and auditable \cite{vaswani2017attention,brown2020language,chen2021evaluating}. Methods for interleaving reasoning with actions and self-supervised API use further strengthen this paradigm, making it feasible to connect language interfaces to domain libraries in a principled way \cite{yao2022react,schick2023toolformer}.

When applied naively to EO imagery, general LLM/VLM systems struggle: they can describe scenes yet fail on quantitative measurement, spatial logic, and consistent use of geospatial metadata. Early multimodal work in the domain (e.g., RSVQA) highlights the need for EO-aware reasoning that spans multiple sensors, resolutions, and coordinate systems rather than surface-level captioning \cite{lobry2020rsvqa}. At the same time, modular program-synthesis frameworks (e.g., ViperGPT~\cite{suris2023vipergpt}) demonstrate that composing specialist vision modules via generated code yields transparency, composability, and reproducibility properties that are equally valuable for Earth Observation workflows, where multiple perception models and geospatial operations must be combined consistently. 

We build on these trends and adopt a code-first design. Our proposed Interpretable Code-based assistant for Earth Observation (\textbf{IC--EO}) compiles natural-language requests into executable, auditable Python code that orchestrates EO tools under an explicit, verifiable plan. Rather than directly providing an answer to a query, the framework operates over a unified Tool API with standardized I/O for scene classification, semantic segmentation, object detection with oriented bounding boxes, spectral indices, and basic geospatial operations such as reprojection, tiling, and area computation. A lightweight controller conditions on sensor and band metadata, selects tools, and validates outputs to ensure consistent spatial processing across RGB and multispectral inputs.% Concretely, the API includes representative EO backbones and operators, DOFA for classification and segmentation~\cite{xiong2024neural}, Prithvi-EO-2.0 for burn-scar segmentation~\cite{szwarcman2024prithvi}, YOLOv11 for oriented object detection~\cite{yolo11_ultralytics}, and standard spectral indices such as NDVI.

To understand the performances of our solution, and compare it to Vision Language Model (VLM)-based approaches, we propose a new evaluation methodology at three complementary levels:
model level to evaluate the performances on public EO benchmarks for classical EO tasks; code-generation level to understand the capacities of a LLM to generate valid python code; and at the task level in which we evaluate the outputs of the framework on two realistic use cases (land-composition mapping and post-wildfire damage assessment) and compare it against strong general-purpose VLM baselines.

After reviewing the state of the art, we detail the method in Section~\ref{sec:method}. Section~\ref{sec:experiments} reports experiments and results at the three levels (model, agent, task). Finally, section~\ref{sec:conclusion} concludes this study by expressing potential implications for EO practice.

\begin{figure*}[!t]
\centering
\includegraphics[width=\textwidth]{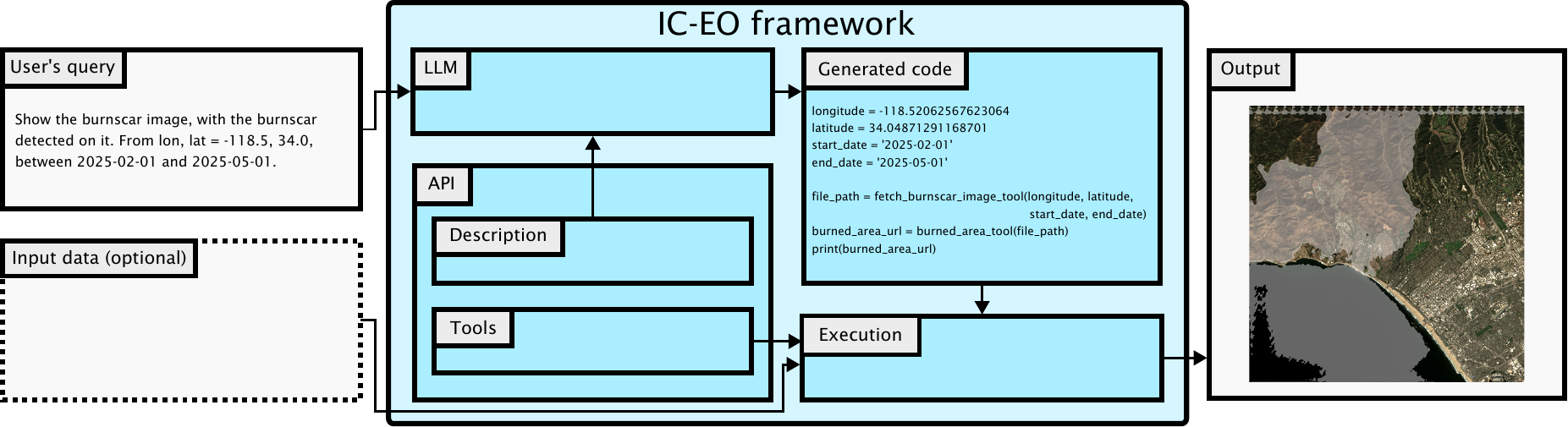}
\caption{\textbf{IC--EO framework overview} From a user's natural language query, optionally with additional input data (such as images), the IC-EO framework produces a freely defined output, e.g. images or natural language. The user's query is first transformed with a LLM to an interpretable Python code which leverages an API or classical and recent EO models. This code can be executed to obtain the answer to the query. }
\label{fig:system-overview}
\end{figure*}

\section{Related Work}
\label{sec:related}

\relpar{Earth Observation Analysis}
While there are more EO data produced than ever, large-scale analysis is now a challenge. Many recent efforts have been done to facilitate this analysis. 
In terms of availability of the data, cloud platforms such as \textit{Google Earth Engine} \cite{gorelick2017earthengine} and the \textit{Microsoft Planetary Computer} \cite{microsoft2022planetary} host petabyte-scale imagery alongside compute resources, enabling users to search, retrieve, and process data directly in the cloud. 
At the modeling layer, open-source libraries like \textit{TorchGeo} \cite{stewart2022torchgeo} provide standardized datasets and pretrained models for many EO tasks, such as land-cover classification, segmentation, and change detection. 
Recent foundation models such as \textit{DOFA} \cite{xiong2024neural} or \textit{RAMEN}~\cite{houdre2025ramen} extend EO understanding to cross-sensor and cross-task reasoning, while benchmarks like \textit{GEO-Bench} \cite{lacoste2023geobench} facilitate consistent evaluation and transfer learning. 
Building on these advances, LLMs and VLMs have been explored for EO interpretation, enabling natural-language interaction with remote-sensing data through tasks such as captioning, description, and visual question answering (VQA). 
The \textit{RSVQA} benchmark \cite{lobry2020rsvqa} introduced the visual question answering (VQA) paradigm to remote sensing, framing EO understanding as natural-language querying of satellite images.
Subsequent work expanded these capabilities: \textit{GeoChat}~\cite{kuckreja2024geochat} and \textit{RS-GPT}~\cite{hu2025rsgpt} enabled conversational analysis of EO scenes, while \textit{ChangeCLIP}~\cite{dong2024changeclip} and \textit{SkyEye-GPT}~\cite{zhan2025skyeyegpt} introduced temporal and video-based reasoning over multi-date imagery. Recently, VLMs have also been proposed to query multi-modal EO data~\cite{boussaid2025visual,tosato2025sar}.
Despite these advances, current models remain perception-oriented: they can describe or classify scenes but cannot autonomously retrieve data, compose preprocessing steps, or execute multi-stage geospatial pipelines. 
Moreover, recent surveys highlight persistent reliability issues in both LLMs and VLMs, including hallucination, factual inconsistency, and spatial misalignment~\cite{liu2024surveyvlm}. Finally, supervising large architectures over EO datasets, have led to a strong presence of biases, which can lower the trust one can have in the outputs of these models~\cite{chappuis2025evaluating}. While models with interpretable bottlenecks~\cite{chappuis2022prompt,chappuis2025pan,tosato2025checkmate} can provide an answer to these biases, they are limited in reasoning capabilities.
These limitations underscore the need for systems that can reason about, compose, and execute geospatial operations autonomously through verifiable computation. 
Our work addresses this by embedding state-of-the-art EO models as callable tools within executable, auditable workflows.

\textbf{LLMs and Code-Based Reasoning}
Recent advances in LLMs make it possible to solve complex visual and analytical tasks by generating executable code rather than free-form text. 
\textit{ViperGPT} \cite{suris2023vipergpt} demonstrated this paradigm, turning visual reasoning into Python program synthesis and achieving strong results across visual question-answering benchmarks. 
\textit{ReAct} \cite{yao2022react} and \textit{Toolformer} \cite{schick2023toolformer} extended it by coupling reasoning with tool use, allowing models to plan, call functions, and verify intermediate outcomes in a step-by-step manner. 
Together, these works establish code-first reasoning as a powerful framework that improves interpretability, composability, and reproducibility. 
In parallel, analyses of modular reasoning strategies \cite{khandelwal2023modular} distinguish between explicit code composition where the model writes and executes programs and natural-language decomposition where reasoning proceeds through successive prompting. 
This comparison directly informs our design choice: we adopt program synthesis as a reliable interface for composing domain tools and inspecting intermediate results. 
Applied to EO, this approach aligns naturally with the field’s workflow structure, where analyses require chaining operations that are difficult to specify in text but straightforward to implement and verify through generated code.

\textbf{EO Agent Systems}
Recent works have explored using LLMs as controllers for geospatial tools, seeking to automate EO analysis through natural-language interfaces. 
Early systems such as \textit{RS-Agent} \cite{xu2024rsagent} and \textit{Change-Agent} \cite{liu2024changeagent} coupled pretrained EO models with LLM reasoning to perform scene understanding and change detection, yet remained limited to static imagery without retrieval or spatial validation. 
The \textit{UnivEarth} benchmark \cite{kao2025univearth} systematically evaluated open LLMs for EO code generation in Google Earth Engine, reporting only 33\% task accuracy and over 58\% execution failures. Another work~\cite{gruening2024benchmarking} introduced one of the first dedicated benchmarks for evaluating large language models on EO tasks, establishing an evaluation framework that highlights current model limitations and motivates validation-aware program synthesis.

Among EO-specific agent architectures, \textit{GeoAgent}~\cite{chen2024geoagent} formalizes analysis as program synthesis, effectively automating vector-based GIS tasks. However, it primarily operates on existing data attributes. More recent systems such as \textit{Earth-Agent}~\cite{feng2025earth} and \textit{GeoLLM-Squad}~\cite{lee2025geollmsquad} broaden this perspective through multi-agent collaboration, but often produce dialogue transcripts rather than executable workflows.

Building on these advances, IC--EO extends the program synthesis paradigm to the \textit{perception} domain. Unlike prior agents that primarily analyze existing GIS layers, IC--EO orchestrates foundation models such as DOFA~\cite{xiong2024neural} and Prithvi~\cite{szwarcman2024prithvi} to generate \textit{new} geospatial layers (e.g., burn scar masks, oriented detections) directly from images. By coupling this capability with a lightweight controller and an API, our system enforces spatial validation and outputs reproducible Python codes, distinguishing it from prior dialogue-based agents.

\section{Methodology}
\label{sec:method}

IC--EO converts natural-language queries into executable Earth Observation (EO) workflows. The framework combines a modular API of selected EO tools with a LLM that plans and writes code. This approach ensures interpretability, reproducibility, and extensibility across tasks. In this section, we detail the API design in~\autoref{ssec:API}. We then describe the LLM-based controller which generates the code to answer a given query in~\autoref{ssec:LLM}.

\subsection{API Design Principles}\label{ssec:API}
The IC--EO API is designed to be modular, maintainable, and aligned with state-of-the-art EO practices. We implement a number of tools, described in~\autoref{sec:experiments} which can be divided in two main categories:
\begin{enumerate}
    \item Data tools: functions which can be used to access EO data, either locally or from external providers;
    \item Model tools: functions which can process EO data to extract information.
\end{enumerate}
Each of these tools is composed of two main components. The first one is the implementation of the tool in itself. The second component is a description which can be used in the prompting mechanism of the LLM to select appropriate tools to answer the input query. This description is systematically structured around pre-defined sections:
\begin{itemize}
    \item A general description of the tool;
    \item A technical description (giving information about the implementation of the tool, inputs and outputs)
    \item In addition, for the model tools:
    \begin{itemize}
        \item A list of supported sensors, including the expected normalization of the inputs;
        \item An example of usage of the tool;
        \item A list of the datasets that were used for training and their taxonomy.
        %\item A section for notes that are determined during experiments (e.g. instructions for using GPU, handling of different data formats, \ldots).
    \end{itemize}
\end{itemize}

% \subsection{Integrated Tools}
% IC--EO integrates a curated set of data utilities and model APIs covering all main EO operations.

% \relpar{Data tools}
% \emph{geeData} and \emph{geeAllData} provide access to Sentinel-1/2 imagery via Google Earth Engine, while \emph{geeUtils} handles band renaming, reprojection, and resampling.

% \relpar{Spectral indices}
% NDVI, SAVI, EVI, NDWI, NDSI, WBI, SR, NWI--1, and NWI--2 offer interpretable, computationally efficient diagnostics.

% \relpar{Models}
% \emph{DOFA} (classification and segmentation); \emph{SAMGeo} (promptable segmentation); \emph{YOLOv11--OBB} (object detection); and \emph{Terratorch} (biomass and burn-scar mapping).

\subsection{LLM-based code generation}\label{ssec:LLM}
A conversational LLM (GPT--4o) serves as the controller that translates user queries into executable Python scripts. Rather than returning predictions directly, it composes workflows using tools defined in the API. The workflow is as follow:
\begin{enumerate}
    \item The user query is combined with the structured tools' descriptions;
    \item The LLM generates executable code invoking IC--EO functions, based on the descriptions;
    \item The script executes in a sandbox and returns figures, masks, or statistics to the frontend;
    \item The generated code is exposed to the user for transparency and reproducibility.
\end{enumerate}
All generated programs run in containerized sandboxes with pinned dependencies, restricted network access, and resource controls. CPU or GPU execution is selected automatically. Logs record tool calls, exceptions, and outputs for full traceability.  
Services are deployed as containers on Microsoft Azure, configured via environment variables for LLM and data-provider access.

% \subsection{Evaluation Protocol}
% IC--EO is evaluated at three complementary levels: code quality, tool accuracy, and end-to-end reliability detailed in Section~\ref{sec:experiments}.

\section{Experiments}
\label{sec:experiments}

The experimental evaluation of IC--EO is designed to assess the full system protocol: how the assistant integrates EO models, helper tools, and a code-generating controller to produce valid, interpretable answers from natural-language queries. We first describe the datasets and tools used in ~\autoref{sec:data_tools}. We then outline the controller setup in~\autoref{ssec:exp:controller}. Finally, we present the three-level evaluation protocol encompassing tool, agent, and task performance in~\autoref{sec:protocol} and the results in~\autoref{sec:results}.

\subsection{Definition of the API}
\label{sec:data_tools}

In this section, we describe the tools implemented in the IC--EO framework. In addition to tools allowing to access data, we implemented tools for classical computer vision tasks, such as image classification, semantic segmentation, promptable segmentation and object detection. Moreover, we have included classical EO indices (e.g. NDVI). Finally, we demonstrate how the proposed framework can be extend by including a specialized tool for mapping burn scars.\\

\noindent\textbf{Scene classification}
We use a classifier based on the DOFA foundation model \cite{xiong2024neural}. DOFA was selected as it allows to use a single encoder for a variety of input sensors. Moreover, as a foundation model, it can be adapted to other tasks. We train classification heads in a linear-probing setup for three datasets:
\begin{itemize}
  \item \textbf{RESISC45}~\cite{cheng2017resisc45}: 31{,}500 RGB images (of size 256$\times$256 pixels) from 45 balanced scene classes covering natural and human-made environments.
  \item \textbf{EuroSAT}~\cite{helber2019eurosat}: 27{,}000 Sentinel-2 patches (of size 64$\times$64 pixels, 13 bands) labeled into 10 land-cover classes.
  \item \textbf{BigEarthNet}~\cite{sumbul2019bigearthnet}: over 590{,}000 Sentinel-2 patches (of size 120$\times$120 pixels for 10m bands, 13 bands) with 43 multi-label CLC-derived classes, forming a large and imbalanced benchmark representative of real-world EO data.
\end{itemize}

To train the classification heads, the encoder is frozen and only a linear classification layer on top of the encoder features is trained. This linear layer is trained for 50 epochs with the LARS optimizer and cosine learning-rate decay. Data augmentation consists of crops with scale in $[0.8, 1.0]$ resized to 224$\times$224 pixels and horizontal flips. We use cross-entropy loss for RESISC45 and EuroSAT, and a multi-label soft-margin loss for BigEarthNet. Learning rates are selected per dataset from the grid $\{0.5, 1.0, 10, 20\}$, and batch sizes are chosen per dataset to balance performance and hardware constraints.\\

\noindent\textbf{Semantic segmentation}
For semantic segmentation, we re-use a DOFA encoder with segmentation heads following the UPerNet architecture~\cite{xiao2018unified}. Two segmentation heads are trained based on the following datasets:
\begin{itemize}
  \item \textbf{Chesapeake}~\cite{robinson2019large}: 1~m aerial imagery over the U.S.\ Chesapeake Bay. The annotation cover seven land-cover classes: water, tree canopy, vegetation, impervious road, impervious other, bare soil and  buildings.
  \item \textbf{FLAIR--2}~\cite{garioud2023flair}: 20~cm aerial imagery over France. Each pixel is annotated among 13 classes: building, pervious surface, impervious surface, bare soil, water, coniferous, deciduous, brushwood, vineyard, herbaceous vegetation, agricultural land, plowed land and "others".
\end{itemize}

The features outputed by the frozen DOFA encoder are transformed into a feature pyramid with 512 channels at four scales (4, 2, 1, 0.5). A UPerNet~\cite{xiao2018unified} segmentation head operates on this pyramid and is the only part trained. The UPerNet head is trained for 20 epochs using the AdamW optimizer, batch size 64, and an initial learning rate of 0.005 with cosine decay. Data augmentation includes center crop, random rotation, and random horizontal and vertical flips; images are normalized using dataset-specific mean and standard deviation. We report mean Intersection-over-Union (mIoU) as the main evaluation metric.

Although other EO foundation models outperform DOFA on some individual datasets, we use DOFA~\cite{xiong2024neural} as the common backbone for our scene classification and semantic segmentation experiments on the datasets above. This simplifies the tools design and avoids retraining or maintaining separate models for each task.\\

\noindent\textbf{Prompt-based segmentation}
\textbf{SAMGeo}~\cite{wu2023samgeo} is integrated as a zero-shot segmentation module providing automatic, point-, box-, and text-driven masks. It is used without fine-tuning and evaluated qualitatively to verify that predicted masks follow expected spatial structures (e.g., vegetation, buildings, water). SAMGeo is particularly useful when the desired taxonomy is not available in our DOFA-based segmentation heads, for example for user-defined land-cover categories or application-specific object classes.\\

\noindent\textbf{Object detection}
We selected a YOLOv11--OBB~\cite{yolo11_ultralytics} model trained on the \textbf{NWPU VHR--10} dataset~\cite{cheng2014nwpu}, which contains ten oriented object categories in very-high-resolution aerial imagery. We use the pretrained weights and configuration released by the authors and do not perform any additional fine-tuning or hyperparameter changes.\\

\noindent\textbf{Burn-scar segmentation}
We use the Prithvi--EO--2.0 foundation model ~\cite{szwarcman2024prithvi} probed on the \textbf{HLS Burn Scars} benchmark, which provides temporal burn-extent mapping from Harmonized Landsat–Sentinel data. We directly use the pretrained checkpoint and settings provided by the authors, without any further fine-tuning.\\

\noindent\textbf{Spectral indices}
We included a library of classical EO indices which can be used on multi-spectral data:\\
\begin{itemize}
  \item \textbf{NDVI (Normalized Difference Vegetation Index)}: 
  Measures vegetation using red (R) and near-infrared (NIR) reflectance:
  \[
    \mathrm{NDVI} = \frac{\mathrm{NIR} - \mathrm{R}}{\mathrm{NIR} + \mathrm{R}}
  \]

  \item \textbf{SAVI (Soil Adjusted Vegetation Index)}: 
  Reduces soil background effects with a soil brightness correction factor \(L\), set at its default value (0.5):
  \[
    \mathrm{SAVI} = \left(\frac{\mathrm{NIR} - \mathrm{R}}{\mathrm{NIR} + \mathrm{R} + L}\right)(1 + L)
  \]

  \item \textbf{EVI (Enhanced Vegetation Index)}: 
  Enhances sensitivity in high biomass regions by including aerosol resistance coefficients ($C_1 = 6$ and $C_2 = 7.5$) and a gain factor $G$ set at 2.5:
  \[
    \mathrm{EVI} = G \cdot \frac{\mathrm{NIR} - \mathrm{R}}{\mathrm{NIR} + C_1 \cdot \mathrm{R} - C_2 \cdot \mathrm{BLUE} + L}
  \]
  
  \item \textbf{NDWI (Normalized Difference Water Index)}: 
  Highlights open water features and is sensitive to changes in liquid water content using green (G) and near-infrared bands.
  \[
    \mathrm{NDWI} = \frac{\mathrm{G} - \mathrm{NIR}}{\mathrm{G} + \mathrm{NIR}}
  \]
  
    \item \textbf{WBI (Water Band Index)}: 
  Indicates canopy water content using reflectance ratio between two near-infrared wavelengths (at 900nm and 970 nm).
  \[
    \mathrm{WBI} = \frac{\mathrm{NIR}_{900}}{\mathrm{NIR}_{970}}
  \]

  \item \textbf{NDSI (Normalized Difference Snow Index)}: 
  Differentiates snow from clouds and other bright surfaces using green and shortwave infrared reflectance (SWIR).
  \[
    \mathrm{NDSI} = \frac{\mathrm{G} - \mathrm{SWIR}}{\mathrm{G} + \mathrm{SWIR}}
  \]

  \item \textbf{SR (Simple Ratio)}: 
  Basic vegetation index defined as the ratio of near-infrared to red reflectance.
  \[
    \mathrm{SR} = \frac{\mathrm{NIR}}{\mathrm{R}}
  \]

  \item \textbf{NWI-1 and NWI-2 (Normalized Water Indices)}: 
  Moisture-related index using near-infrared and first or second shortwave infrared band.
  \[
    \mathrm{NWI}\text{-}1 = \frac{\mathrm{NIR} - \mathrm{SWIR}_1}{\mathrm{NIR} + \mathrm{SWIR}_1} \qquad \mathrm{NWI}\text{-}2 = \frac{\mathrm{NIR} - \mathrm{SWIR}_2}{\mathrm{NIR} + \mathrm{SWIR}_2}
  \]
\end{itemize}

\noindent\textbf{Data tools}
We implemented a tool which interfaces the Google Earth Engine API~\cite{gorelick2017earthengine} to access Sentinel 1 (Synthetic Aperture Radar) and 2 (Multi-spectral imagery). Additionally, we include tools for standard preprocessing, mosaicking, and reprojection of geo-coded data. These tools ensure cohesive and reproducible end-to-end workflows.

\subsection{Definition of the Controller}\label{ssec:exp:controller}

The controller of IC--EO is a large language model (LLM) responsible for translating natural-language instructions into executable code. We use \textbf{GPT--4o} as the primary controller due to its balanced trade-off between coding accuracy, speed, and cost. In preliminary comparisons, GPT--4o showed high reliability in generating syntactically valid and semantically correct tool calls, while remaining responsive enough for interactive experimentation. The controller operates strictly under a code-only contract: it must output runnable Python code without any prose, markdown, or explanations. The system prompt explicitly enforces this format and requires that the program always print a final result.

Tool specifications including function names, argument types, and return values are provided to the model as structured text, described in~\autoref{ssec:API}. This explicit schema narrows the generation space and reduces invalid function calls, a known source of hallucination in tool-augmented LLMs. In this setup, the LLM itself performs both planning (deciding which tools to use and in what order) and code synthesis (writing the corresponding Python code), unifying reasoning and execution planning in a single step. 

\subsection{Execution environment and guardrails}

Generated code is executed in a sandboxed runtime with restricted permissions and resource limits. Each run is fully logged capturing the generated code, tool calls, outputs to the user, and exceptions so that any result can be reproduced or audited. The system includes a chat-based frontend and a FastAPI backend. The backend retrieves the tool registry, invokes the LLM for code generation, and executes the program in a sandboxed runtime. GPU-intensive tasks are dispatched to NVIDIA T4 instances automatically.

This design achieves a balance between flexibility and safety: the controller can freely compose available tools to build pipelines, but within a controlled environment that guarantees transparency and traceability. %In practice, this configuration yielded an execution success rate of 87\% across 600 test prompts, confirming both the reliability of GPT--4o as a code generator and the robustness of the execution harness.

\subsection{Evaluation protocol}
\label{sec:protocol}

The IC--EO framework is built following a modular approach. Hence, we propose an evaluation protocol at three levels:
\begin{itemize}
    \item \textbf{Final answer} We first evaluate the correctness of the answer for a number of queries. For each of our test query, we produce a ground truth annotation. We produced two datasets covering two scenarios:
    \begin{enumerate}
        \item land-cover composition: based on an input image, we ask questions such as \emph{Is there snow in this image?} or \emph{Are there some greenhouses in the area?}. We built 600 questions for this test set.
        \item post-wildfire assessment: based on an input image and json data simulating insurance files, we ask questions such as \emph{What was the total damage in terms of dollar amount during the fire at lon, lat = -118.52062, 34.04871 that happened in January of 2025}. For this test set, we built 12 questions. Moreover, we added 12 simpler questions covering the presence of objects in the image, such as \emph{How many vehicles?}. 
    \end{enumerate} 
    \item \textbf{Tool-level} Each EO model is tested independently on its modality-native benchmark using standard splits. We report top-1 accuracy for classification (RESISC45, EuroSAT, BigEarthNet), mean Intersection-over-Union (mIoU) for segmentation (Chesapeake, FLAIR--2), mean Average Precision at 50\% rotated IoU (mAP@50) for detection (NWPU VHR--10), and IoU for burn-scar segmentation (HLS Burn Scars). Zero-shot components such as SAMGeo and spectral indices are evaluated qualitatively.
    \item \textbf{LLM-level} We assess the assistant’s ability to generate valid (i.e. executable), hallucination-free code under tool constraints. The main metrics are the \emph{Execution Success Rate},the fraction of generated scripts that execute without runtime errors in the sandbox and the \emph{Code Validity Rate}, defined as the proportion of syntactically and semantically correct tool calls.
\end{itemize}

% ----------------------
% RESULTS 
% ----------------------
\subsection{Results}\label{sec:results}

\subsubsection*{Evaluation of the final answer}
The results on our two proposed datasets are shown in \autoref{tab:l2-scenario-combined}. On both of these datasets, we can observe that IC--EO obtains higher performances than vanilla VLMs. This demonstrates the advantage of leveraging well-defined computer vision algorithms for EO information extraction. Notably, for high-level of reasoning tasks, exemplified by the Burn scars case study, none of the VLMs managed to answer correctly to the queries. This demonstrates that our modular approach is relevant. In addition to better performing model, our proposed framework allows to examine the reasoning that was used to produce an answer. We present two runs showing (i) a successful detection and (ii) a hardware-related failure that does not stem from code generation.
\begin{table}[H]
\centering
\footnotesize
\begin{tabular}{l c c c}
\toprule
System & Land cover & \multicolumn{2}{c}{Post-wildfire assessment} \\
\cmidrule(lr){3-4}
 & Accuracy (\%) & Burn scars accuracy (\%) & Objects accuracy (\%) \\
\midrule
IC--EO (ours)       & \textbf{64.2} & \textbf{50.0} & \textbf{58.0} \\
GPT--4o baseline    & 51.7          & 0.0           & 25.0          \\
LLaVA baseline      & 42.3          & 0.0           & 8.0           \\
\bottomrule
\end{tabular}
\caption{Accuracy of the IC--EO system compared with two VLM baselines on land-cover and post-wildfire assessment tasks.}
\label{tab:l2-scenario-combined}
\end{table}

\begin{figure}
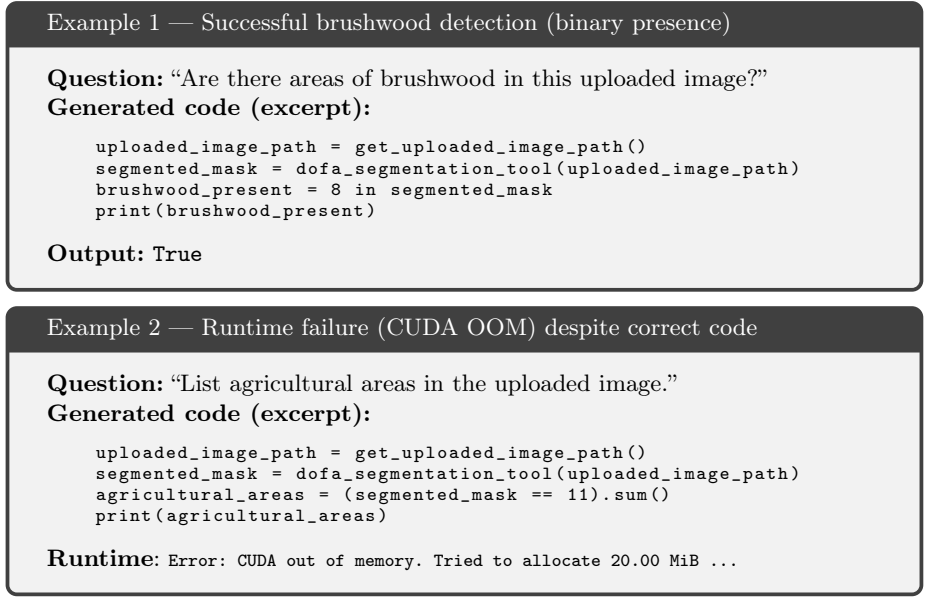

    \centering
    \begin{tcolorbox}[title={Example 1 — Successful brushwood detection (binary presence)}]
    \textbf{Question:} ``Are there areas of brushwood in this uploaded image?''
    
    \textbf{Generated code (excerpt):}
    \begin{lstlisting}[language=Python,basicstyle=\ttfamily\scriptsize]
    uploaded_image_path = get_uploaded_image_path()
    segmented_mask = dofa_segmentation_tool(uploaded_image_path)
    brushwood_present = 8 in segmented_mask
    print(brushwood_present)
    \end{lstlisting}
    
    \textbf{Output:} \texttt{True}
    \end{tcolorbox}
    
    \begin{tcolorbox}[title={Example 2 — Runtime failure (CUDA OOM) despite correct code}]
    \textbf{Question:} ``List agricultural areas in the uploaded image.''
    
    \textbf{Generated code (excerpt):}
    \begin{lstlisting}[language=Python,basicstyle=\ttfamily\scriptsize]
    uploaded_image_path = get_uploaded_image_path()
    segmented_mask = dofa_segmentation_tool(uploaded_image_path)
    agricultural_areas = (segmented_mask == 11).sum()
    print(agricultural_areas)
    \end{lstlisting}
    
    \textbf{Runtime}: {\scriptsize\texttt{Error: CUDA out of memory. Tried to allocate 20.00 MiB ...}}
    \end{tcolorbox}
    \caption{Two short runs of IC--EO.}
    \label{fig:my_label}
\end{figure}

\subsubsection*{Evaluation of the tools}
We evaluate the tools defined in the IC--EO API on common EO benchmarks to establish a performance baseline (Table~\ref{tab:l1}). Note that while we use existing models from the literature, these results confirm that our tools are close to state-of-the-art accuracy on their modality-specific domains.

\begin{table}[H]
\centering
\small
\setlength{\tabcolsep}{10pt}
\renewcommand{\arraystretch}{1.12}
\begin{tabular}{l l c c}
\toprule
Task & Dataset & Metric & Score \\
\midrule
Classification & EuroSAT       & Top-1 (\%)     & \textbf{94.0} \\
               & BigEarthNet   & Top-1 (\%)     & \textbf{68.0} \\
               & RESISC45      & Top-1 (\%)     & \textbf{97.0} \\
\midrule
Segmentation   & Chesapeake & mIoU          & \textbf{64} \\
               & FLAIR--2  & mIoU          & \textbf{50} \\
\midrule
Detection      & NWPU VHR--10   & mAP@50 (\%)   & \textbf{80.9} \\
Burn-scar      & HLS Burn Scars & IoU (burned)  & \textbf{87.5} \\
\bottomrule
\end{tabular}
\caption{Tool-level global metrics (L1).}
\label{tab:l1}
\end{table}

\subsubsection*{Evaluation at the LLM level}
Finally, we show the performances of the LLM as a code generator in \autoref{tab:ign-generated-code-outcomes}. These results shows that the adopted structure for the API, described in~\autoref{ssec:API} is well understood by the LLM. Note that among the 13\% of codes that were not able to execute, a majority of failures is due to execution-time constraints (e.g., GPU memory), rather than incorrect code or tool choice. This experiment shows that there is a large gap between the percentage of code that is well generated by the LLM, and the final accuracy of the IC--EO framework (64.2\%).
\begin{table}[t]
\centering
\small
\setlength{\tabcolsep}{10pt}
\renewcommand{\arraystretch}{1.1}
\begin{tabular}{lc}
\toprule
\textbf{Model/System} & \% Valid code  \\
\midrule
IC--EO Agent (ours) & \textbf{87.0}  \\
\bottomrule
\end{tabular}
\caption{Percentage of executable codes generated and executed by the LLM on the land cover test set.}
\label{tab:ign-generated-code-outcomes}
\end{table}

\section{Conclusion}
\label{sec:conclusion}

Earth observation (EO) analysis requires the integration of data access, geospatial preprocessing, and task-specific models, yet these components are often fragmented across scripts and specialized tools. IC--EO addresses this gap through a modular, code-generating agent that translates natural-language queries into executable programs orchestrating EO-specific functions. The framework adopts a ViperGPT-style design emphasizing transparency, deterministic spatial logic, and reproducibility, combining a code-only controller with structured tool specifications, a lightweight execution harness, and a curated palette of models including DOFA for classification and segmentation, YOLOv11--OBB for oriented-object detection, Prithvi--EO--2.0 for burn-scar mapping, and SAMGeo for prompt-based segmentation, alongside utilities for Sentinel-2 retrieval and preprocessing. With this approach, we ensure three key properties:

\noindent\textbf{Interpretability.} The LLM generates executable Python code specifying the sequence of functions, preprocessing steps, and models used allowing full traceability of each result.

\noindent\textbf{Reproducibility.} Prompt structures and tool interfaces are standardized so that identical queries yield identical code and outputs. Evaluation relies on fixed datasets and widely-used metrics such as accuracy, Intersection over Union (IoU), and mean Average Precision (mAP).

\noindent\textbf{Extensibility.} Each tool—data access, spectral index, or model is wrapped as a self-contained Python module that registers automatically in a central registry. New tools or datasets can be added without modifying the core framework. With this approach, the proposed framework can be extended to support new tasks, and benefit from new, better performing models.

% \noindent\textbf{Automation.} Two automated mechanisms maintain consistency: 
% (i) an automatic documentation pipeline that extracts and structures tool metadata for the LLM, and 
% (ii) a deterministic model routing mechanism that inspects the input tensor's channel count and spatial resolution to automatically route queries to the appropriate backbone.\\

Evaluation across two proposed test sets demonstrate consistent improvements from explicit tool orchestration compared to end-to-end vision–language inference. Qualitative analysis shows that IC--EO performs most reliably on concrete, computation-defined tasks such as classification, counting, and area estimation, while comparative or temporal prompts reveal gaps related to default change analytics and summarization. Most observed failures arise from toolchain constraints, such as memory limits on high-resolution imagery or incomplete preprocessing, rather than from the controller’s planning logic. Overall, the results suggest that progress in EO assistants will be driven less by monolithic multimodal models and more by structured composition: stronger base models, explicit geospatial reasoning, dependable execution for large-scale data, and controllers that generate transparent, reproducible code for scientific analysis.

\subsubsection{Acknowledgements} This research has been carried under the ESA Contract No. 4000144684/24/I-DT-bgh.

\bibliographystyle{splncs04}
\bibliography{refs}

\end{document}